\documentclass[a4paper]{article}
\usepackage{hyperref}
\usepackage[colorinlistoftodos,prependcaption,textsize=tiny, linecolor=green,     backgroundcolor=white, disable]{todonotes}

\newcommand{\textcite}{\cite}
\usepackage{INTERSPEECH2020}

\usepackage{amsmath}
\usepackage{subcaption}

\newcommand{\eqdef}{\overset{\mathrm{def}}{=\joinrel=}}
\graphicspath{ {./img/} }

\DeclareMathOperator*{\argmin}{arg\,min}

\title{Zero-shot Multi-Domain Dialog State Tracking Using Descriptive Rules}
\name{Edgar Altszyler$^{1,2,\dagger}$, Pablo Brusco$^{1,2,\dagger}$, Nikoletta Basiou$^3$, John Byrnes$^3$, Dimitra Vergyri$^3$}
\address{
  $^1$Departamento de Computación, FCEyN, Universidad de Buenos Aires, Argentina\\
  $^2$Instituto de Investigación en Ciencias de la Computación, CONICET-UBA, Argentina\\
  $^3$SRI International, USA
  }
\email{\{ealtszyler, pbrusco\}@dc.uba.ar,\{nikoletta.basiou, john.byrnes, dimitra.vergyri\}@sri.com\\
  $\dagger$ These authors contributed equally to this work. This work was performed while the authors were still at SRI.}

\begin{document}

\maketitle
\begin{abstract}
 In this work, we present a framework for incorporating descriptive logical rules in state-of-the-art neural networks, enabling them to learn how to handle unseen labels without the introduction of any new training data. The rules are integrated into existing networks without modifying their architecture, through an additional term in the network's loss function that penalizes states of the network that do not obey the designed rules.
 As a case of study, the framework is applied to an existing neural-based \textit{Dialog State Tracker}. Our experiments demonstrate that the inclusion of logical rules allows the prediction of unseen labels, without deteriorating the predictive capacity of the original system.
\end{abstract}
\noindent\textbf{Index Terms}: zero-shot learning, differentiable logic, neural networks, dialog state tracker, dialog systems

\section{Introduction}
When deploying machine-learning-based systems, it is common for users to detect problems related to functionalities that do not meet the expected requirements. In particular, in dialog systems, problems arise when for a certain input, the model makes a prediction that is different from the user's inferred decision. This is often due to the model structure and to the inherent characteristics of the dataset used to train the models. 
In the same direction, new user requirements for the system may require outputing \textbf{unseen labels} (not present in training data). In such situations, a typical solution consists of the collection of new annotated data aligned with the expected functionality. However, collecting new data every time such a need arises is an expensive and time-consuming effort, therefore, an alternative approach is desired. 

In this work, we propose a solution to the above mentioned problems by incorporating \textbf{descriptive logical rules} into learned neural network models. These rules are designed by domain experts and can influence the system output, enabling also the prediction of unseen labels.
Similar to other works \cite{Hu2016,sikka2020deep, Xu2018, fischer2019dl2, li2019augmenting}, we use differentiable first-order logic (FOL) that has proven useful for integrating knowledge into a neural-symbolic system. 


We apply our logic rules framework to Dialog State Tracking, a challenging and complex task in the field of Dialog Systems. More specifically, we extend the Multi-Domain Neural Belief State Tracker (MDNBT), proposed in \cite{ramadan2018large} and recently incorporated as one of the state of the art dialog state trackers in ConvLab, an open-source multidomain end-to-end dialog system platform released under the Dialog State Tracker Challenge (DSTC8) \cite{lee2019convlab}. 

%
%

The main contributions of our work are the following: a) we enhance a Neural-based Dialog State Tracker with logic rules, without degrading the performance of the base system, b) we show that the addition of the logic rules allow the predictions of unseen labels which can be very useful in the case of unlabeled or partially labeled data.


\section{Related Work}
\label{sec:related_work}

Due to the increasing popularity of neural network models for supervised learning, there is a growing body of material related to the inclusion of structural knowledge as a tool of biasing certain models decisions and as a way to mitigate the uninterpretability of results. One way to introduce this knowledge is to integrate logical rules through the use of FOL --- a declarative language that can represent high-level knowledge \cite[inter allia]{Hu2016, sikka2020deep, Xu2018, fischer2019dl2, li2019augmenting, Zhang2019, Marra2019, Chen2019,van2020analyzing,marra2020relational}. In the majority of prior works, different types of posterior regularization terms are implemented to affect the optimization process. For example, in a seminal work by Hu et al. \cite{Hu2016}, the authors propose a general teacher-student framework approach in which the model simultaneously learns from labeled data and logical rules through an iterative process that shifts the parameters of CNN and RNN networks in the tasks of sentiment classification and named entity recognition. Although these works generally apply rules as functions of the network inputs and outputs, there are some that can predicate over the internal values of the network's neurons \cite{li2019augmenting}. These rules can be included in the existing neural network architectures to guide the training and predictions without additional learning parameters \cite{Hu2016, van2020analyzing}. 
There are also works that use these techniques in a semi-supervised setting. For example, in \textcite{Hu2016, Xu2018} it is shown that the rules can be applied over unlabeled data. However, to our knowledge, this type of framework has not been applied to the fully-unsupervised problem of unseen-labels prediction (\textit{zero-shot learning}). 


\section{Our approach}
\label{sec:methods}

The proposed framework consists of the addition of a plug-in component into an existing computational graph with \textbf{no extra learning parameters}. A rules-dependent loss term is introduced into the system's loss function as a way of integrating rules to the learning process and as a way of allowing the use of off-the-shelf optimizers. By including rules as part of the training process, we generate extra cost when the network does not satisfy a certain rule. 


\subsection{Neural Belief State Tracker}

A \textit{Dialog State Tracker (DST)} is a key component in task-based spoken dialog systems. It models the user’s intent at any point of an ongoing conversation \cite{young2010} which is then used by the downstream dialog management component to choose the next system response. DST models estimate the \textit{belief state}, which is the system's internal probability distribution over possible dialog states, by taking into account the user goals at every turn as extracted by a Spoken Language Understanding (SLU) component. The dialog states are defined by a domain-specific ontology that lists the slot-value pairs that describe the constraints the users can express (e.g. price range-expensive, price range-cheap, area-west, area-east, etc) \cite{mrkvsic2018fully}. 

In our work, we use the Multi-domain Neural Belief State Tracker (MDNBT) which jointly identifies the domain and tracks the belief states corresponding to that domain by utilizing the semantic similarity between dialog utterances and ontology terms \cite{ramadan2018large}. MDNBT is implemented as multi-layer networks with Bi-LSTMs to model the user and system utterance and with RNNs with a memory cell to model the flow of the conversation.

For our experiments, we used the MultiWOZ 2.0 dataset \cite{budzianowski2018mwoz,ramadan2018large}. This dataset contains 2480 single-domain dialogues and 7375 multiple-domain dialogues, in which at least two domains are involved throughout each conversation. The ontology contains a total of 663 domain-slot-values triples, distributed across 27 slots in 5 domains (restaurant, hotel, attraction, train, taxi). Here we show an example of two turns inside a conversation with their corresponding state label.
 \begin{enumerate}

 \item \textbf{utterance:} Hi, can you help me find a place to eat on the northside?  
 
 \{\texttt{True state: restaurant-area-north}\}

 \textbf{system:} Yes, I have 15 options, do you have any preferences for the price range?

 \item \textbf{utterance:} Yes, I would like an expensive restaurant

\{\texttt{True state: restaurant-area-north, restaurant-pricerange-expensive}\}

 \textbf{system:} There are 2 expensive places, an Italian restaurant and a gastropub.

 \end{enumerate}





\subsection{Rules definition}
\label{sec:rules-def}

Rules are defined as formulas in a relaxation of FOL that represent truth values in a continuous domain in which the satisfaction of rules is a differentiable function that can be maximized to perform learning \cite{sikka2020deep, van2020analyzing}. A formula's truthiness is represented as a real number that indicate the degree of truth or falsity of a relation defined over entities of the system\footnote{In our experiments, we are not interested in FOL functions and quantifiers, and we leave them out of the discussion.}.

We formulated two types of rules for this study. The type $R_1$, which triggers when some specific keyword (in this case \textit{expensive}) is explicitly uttered in a specific domain (in this case \textit{hotel}), and the type $R_2$ in order to preserve belief-state predictions related to the price-range slots throughout the turns for the cases when the user's price-range intent does not change throughout the turns. Examples of the two types of rules for the \emph{HOTEL} domain are given below:

\noindent $R_1$:  
\textbf{IF} the \textbf{user's utterance} \underline{contains a word like} \emph{EXPENSIVE} \textbf{AND} also \underline{contains a word  like} \emph{HOTEL}  \textbf{THEN} the \textbf{prediction}  in the domain \emph{HOTEL}, slot \textit{PRICERANGE} \underline{should be} \textit{EXPENSIVE}

\noindent $R_2$: \textbf{IF} the \textbf{previous prediction} for the domain \textit{HOTEL}, slot \textit{PRICERANGE} \underline{was} \textit{EXPENSIVE} 
\textbf{AND} the user did \textbf{NOT} \textbf{uttered} \underline{a word like} \emph{MODERATE} \textbf{AND} did \textbf{NOT} \textbf{uttered} \underline{a word like} \emph{CHEAP}   
\textbf{THEN} the next \textbf{prediction} in the domain \textit{HOTEL}, slot \textit{PRICERANGE}  \underline{should be} \textit{EXPENSIVE}.

In these examples, underlined words represent \textit{predicates}; bold-uppercase terms represent \textit{logical connectives}; bold-lowercase terms represent nodes in the computational graph on the network (or concepts that are mapped into embeddings such as in the case of the user's utterance); finally, italic-uppercase words refer to the \textit{constants} of our system.

 As we will discuss in further detail in the next sections, rules are implemented in the model's computational graph through the addition of new operations that are applied over existing nodes. When the graph is evaluated for a specific instance, each rule produces a number that indicates the \textit{truthiness} of that rule for the specific instance under evaluation. The loss function will be a function of the \textit{truthiness}, allowing the network to learn the rules.

\subsubsection{Learning mechanism}



Unsatisfied rules generate a cost that the optimizer minimizes in conjunction with the misclassification cost. Our system is trained not only to learn from labels (by minimizing the cross-entropy loss function) but also to learn how to make the set of rules $\mathcal{R}$ as true as possible (a concept called \textit{best satisfiability} as presented in \cite{Donadello2017}). For this, we use the simplest posterior regularization approach, in which the objective function of the rule-based model is the sum of the loss function of the base MDNBT model ($\mathcal{L}_{MDNBT}$) and the rules' loss function ($\mathcal{L}_{rules}$) \cite{van2020analyzing},
\begin{equation}\label{eq:loss_full}
    \mathcal{L(\theta; \mathcal{D}; \mathcal{Y}; \mathcal{R})} = \mathcal{L}_{MDNBT}(\theta;  \mathcal{D}; \mathcal{Y}) + w\; \mathcal{L}_{rules}(\theta;  \mathcal{D}; \mathcal{R}) 
\end{equation}
\noindent where $\theta$ represents the model parameters (weights and biases), $\mathcal{D}$, $\mathcal{Y}$ refers to the dataset and its labels respectively, $\mathcal{R}$ represents the set of rules of our system, $w$ is a weighting hyperparameter that we call \textit{rules' weight}, and $\mathcal{L}_{rules}$ is defined as the sum of the loss of each individual rule:  
\begin{equation}\label{eq:loss_RULES}
    \mathcal{L}_{rules}(\theta;  \mathcal{D}; \mathcal{R})  = \sum_{r \in \mathcal{R}} \mathcal{L}_{r}(\theta;  \mathcal{D}) = 
    \sum_{r \in \mathcal{R}} 1-truthiness_r(\theta;  \mathcal{D})
\end{equation}
Here, each individual rule loss ($\mathcal{L}_{r}$) is defined in terms of the $truthiness_r$, i.e. the degree of truth of the rule.
Finally, the optimization problem consists of finding the optimal weights and biases for the network given the composed loss function,
\begin{equation}\label{eq:optimization_problem}
    \theta^{*} = \argmin_{\theta}  \mathcal{L(\theta; \mathcal{D}; \mathcal{Y}; \mathcal{R})}
\end{equation}
The backpropagation mechanism computes gradients that update all trainable parameters in the network that are reachable from the loss function. For weights and biases to be reachable, the operations that define the loss function have to be fully differentiable. Thus the logical operations that define truthiness of the rules need to be differentiable.

\subsubsection{Formulas and Predicates}
We define two types of formulas: (i) atomic formulas: predicates applied to constants and nodes of the computational graph; and (ii) composed formulas, which are built up from atomic formulas using the Boolean connectives.

Predicates define relations among entities of the neural network. 
For example, in the case of the MDNBT model, we can refer to the embedding representation of a word in the input data; to the actual belief states ($b_{t}$) --- a slot-specific distribution of probabilities estimated by the MDNBT output layer \cite{ramadan2018large};  to previous belief states ($b_{t-1}$); etc. Also, predicates may refer to external values to the computational graph (i.e constants) such as a pre-trained word embedding for the word \textit{HOTEL}. When the computational graph is evaluated, each predicate returns a truthiness value. That is, a value between $0$ and $1$, with $1$ being the highest confidence in the truth of a predicate.


The implementation of a predicate, as opposed to that of a logic connective, strongly depends on the underlying architecture and on the input representations. For example, the predicate \underline{contains a word like} CHEAP can be expressed as the cosine similarity function over nodes of the computational graph: $In_{thresh=t}(utt, \mbox{CHEAP})$. This function may check if there is any word in the utterance whose cosine similarity with the word embedding for \textit{cheap} is greater than threshold $t$.

\subsubsection{Logical operators}
\label{sec:logic}

Logic operators (namely  $\land, \lor, \lnot, \rightarrow{}$) are implemented through the following equations based on \textit{Product} $t$-norm for conjunction,  $s$-norm for disjunction, and the \textit{residuum} of the $t$-norm for the implication (see \cite{Serafini2016} and \cite{van2020analyzing} for further details). 

\begin{align*}
\lnot X  &\eqdef 1-X, \\
X \land Y &\eqdef X \cdot Y, \\
X \lor Y &\eqdef \lnot (\lnot X \land \lnot Y) = X + Y - XY \\
X \rightarrow Y &\eqdef \lnot (X \land \lnot Y) = 1-X(1-Y)
\end{align*}
Having defined all the aforementioned components, we can represent rules in terms of logic formulas. For example, $R_1$ is:

\smallskip

\noindent $R_1 \equiv (In_{thresh=t}(utt, \mbox{EXPENSIVE}) \land \\ 
~~~~~~~~~~~~~~~~~~~ In_{thresh=t}(utt, \mbox{HOTEL})) \rightarrow \\ 
~~~~~~~~~~~~~~~~~ Assert(b_t, \mbox{HOTEL-PRICERANGE-EXPENSIVE})
$

\smallskip

\noindent where $Assert(b_t, \mbox{HOTEL-PRICERANGE-EXPENSIVE})$ is a predicate that returns the model's belief state probability $b_t$ in the index corresponding to the \textit{HOTEL-PRICERANGE-EXPENSIVE} state.

\subsubsection{Antecedent and Consequent learning}
\label{eq:gradient}
When a gradient-based method is used to solve the optimization problem of Eq. \ref{eq:optimization_problem}, the parameters $\theta$ of the network are updated in the opposite direction to the gradient of $\mathcal{L}$. The rule loss term produces an update associated with it.  For example, in the case of a rule with the form $r = X \rightarrow Y$ in which the loss is $\mathcal{L}_{r} = X(1-Y)$ , the update associated to the rule looks like:

\begin{align*}
\Delta_r\theta = - \lambda \big( d_X(\mathcal{L}_{r}) d_{\theta}(X) )+ d_Y(\mathcal{L}_{r}) d_{\theta}(Y) )\big)
\end{align*}
where $\lambda$ is the learning rate and the partial derivatives of the implication are $d_X(\mathcal{L}_{r})= 1-Y \text{ and } d_Y(\mathcal{L}_{r})= -X$.

For example, in the case where this implication is not satisfied ($X=1$ and $Y=0$), the partial derivatives are $d_X(\mathcal{L}_{r})=1$ and $d_Y(\mathcal{L}_{r})=-1$, and the update is,
\begin{equation*}
    \Delta_r\theta = - \lambda \, d_{\theta}(X) + \lambda \, d_{\theta}(Y)
\end{equation*}
The network will update $\theta$ in the direction of growth of Y and in the direction of decrease of X. 
That is, during the learning step the antecedent tends to decrease and the consequent  tends to increase simultaneously. 

Depending on the rule, one may want the antecedent or the consequent learning to be \textit{frozen} (i.e. that the learning process occurs only through one of them). 
This is the case of some \textit{if-type} rules like $R_2$, in which we are not interested in learning how to make the \textit{condition} {\em True}, but we want to make the \textit{then branch} {\em True} (in case the condition is met). For these types of cases, one can use two different alternatives: either implementing predicates using non-derivable functions such as $argmax$; or by programmatically stopping the back-propagation in the corresponding subterms.

\section{Results and Discussion}

The main question we address in this section is: \textit{can we extend a system with a new slot without any additional data and without degrading the existing system?} 

To answer this question, we simulated this scenario by removing all existing annotations for the \textit{PRICERANGE} slot in the training set of the MULTIWOZ dataset.  Next, we built a set of twelve rules designed to learn the \textit{PRICERANGE} slot values. Six of these rules were in the form of $R_1$, and six were in the form of $R_2$ (as described in Section \ref{sec:rules-def}), addressing the six possible combinations of domains (hotel and restaurant) and price ranges (cheap, moderate and expensive).

We compare our models to a \textit{base} MDNBT model that does not contain rules and in which we removed the \textit{PRICERANGE} slot-value pairs from the ontology. In our experiments, we used $100$ Bi-LSTM cells and we trained the models from scratch using the ADAM optimizer \cite{kingma2014} with batch size $64$ for 600 epochs. A dropout rate \cite{srivastava2014dropout} of $50\%$ was used in all the intermediate representations. Also, all the weights were initialized using normal distribution of zero mean and unit variance and biases were initialized to zero. For the rule-based MDNBT we trained models with four different values of the rules' weight, i.e., $w=10, 30, 100, 300$. 

\begin{figure}[t]
    \centering
    \includegraphics[width=\linewidth]{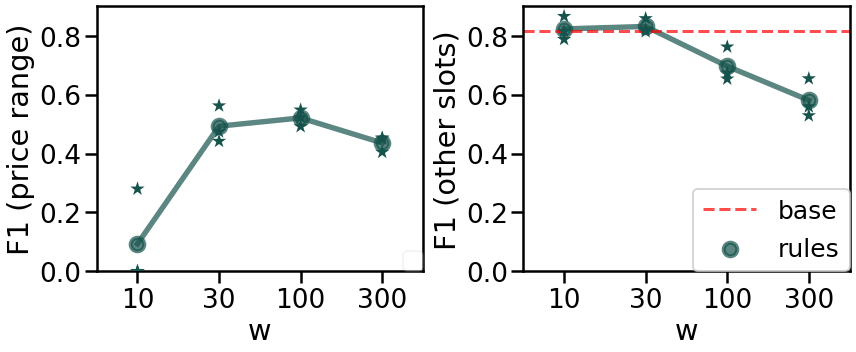}

    \caption{Performance of rules-based MDNBT models for the \textit{PRICERANGE} slots (left), and for the rest of the slots (right). Each star shows the $F_1$ for 3 different runs per weight, and the mean values are shown as circles. The mean $F_1$, in the \textit{PRICERANGE} slots, is 0.093,0.493, 0.521 and 0.436 for the values of $w$ of 10, 30, 100, 300 respectively. While the mean $F_1$ in the other slots is 0.823, 0.697, 0.832, and 0.581 for the $w$ of 10, 30, 100, and 300 respectively. Performance of the \textit{base MDNBT model} is also included in the right plot, which is computed as the mean among 6 runs ($\mu=0.817\pm0.085$).}
    
    \label{fig:performance}
\end{figure}

The F1 performance of the rules-based MDNBT models for the \textit{PRICERANGE} slot and for all the remaining slots are depicted in Fig.~\ref{fig:performance}. F1 is measured by considering the correct and incorrect predictions in each slot of each domain. The performance of the base MDNBT model for the remaining slots is also shown (right plot). 
From the left plot, it is evident that the rule-based MDNBT model learns how to predict \textit{PRICERANGE} slot values without any training data (zero-shot learning). The increase in the performance measured in the \textit{PRICERANGE} slots for different rules' weight values affects the overall performance on the rest of the slots in the ontology depending on the rules' weight value. As can be seen from the right plot, for lower rules' weight values (e.g. $w=10$) in the rule-based MDNBT there is not an appreciable performance drop compared to the base MDNBT ( $0.7\%$ relative decrease), while for large rules' weights (e.g. $w=300$) there is a considerable degradation of performance (i.e., $29\%$ relative decrease). 
A selection of an optimal rules' weight (e.g. $w=30$) can guarantee the optimal trade-off between the performance on unlabeled data (i.e. \textit{PRICERANGE} slot) and the performance on labeled data (e.g. the remaining slots in the ontology).  

Our results do not show a significant performance degradation in the other slots (right panel) for low rules-weight values ($w = 10,30$), since it does not show significant difference with the base model (two-sided $t$-test $p$-val $>$ 0.1 in both $w$ values). However, using $w=100$ or $w=300$ we see a notable decrease in the general performance ($p$-val = 0.06 and $p$-val= 0.004 respectively with two-sided $t$-tests). That is, the system learns how to identify price ranges at the expense of producing unwanted effects in the performance of the rest of the slots.


From the experimental results, we observe that it is possible to integrate rules into an existing system to allow the prediction of unseen labels without degrading the predictive capabilities over the rest of the labels (as it is the case with $w=30$). However, it is necessary to pay special attention to the trade-off that is generated between learning the rules and the degradation of the system. In particular, it is important to notice that the weights depend on the number of times the rules are actually satisfied, the number of rules, and the design properties of the system.

\section{Conclusions}
This paper presented how the addition of descriptive logical rules designed by domain experts can enable neural networks to predict unseen labels without the need for creating new labeled training data. The rules are integrated into an existing neural network without modifying the original architecture. A posterior regularization approach was used to introduce the rules into the learning process, penalizing the objective function when inputs and the internal state of the network do not obey one of the designed rules. 
Our rules-based framework was applied and tested to an existing neural-based \textit{Dialog State Tracker} for Dialog Systems where rules were implemented so that the model learns to identify \textit{PRICERANGE} labels, which were not seen during training. Our experiments showed that the inclusion of logical rules allows the prediction of new labels, without jeopardizing the predictive capacity on the rest of the data. It is finally worth noting that our rules-based solution is independent of the neural network model and thus can be applied to any application (and neural network model) given the formulation of appropriate rules.

\bibliographystyle{IEEEtran}
\bibliography{mybib}

\end{document}